\documentclass[lettersize,journal]{IEEEtran}
\usepackage{amsmath,amsfonts}
\usepackage{algorithmic}
\usepackage{algorithm}
\usepackage{array}
\usepackage[caption=false,font=normalsize,labelfont=sf,textfont=sf]{subfig}
\usepackage{textcomp}
\usepackage{stfloats}
\usepackage{url}
\usepackage{verbatim}
\usepackage{graphicx}
\usepackage{cite}
\usepackage{amsmath}
\usepackage{xcolor}

\usepackage{booktabs}
\usepackage{bbding}
\usepackage{multirow}

\begin{document}

\title{DiffusionTrend: A Minimalist Approach to Virtual Fashion Try-On}

\author{
Wengyi Zhan,~\IEEEmembership{}
Mingbao Lin,~\IEEEmembership{}
Shuicheng Yan,~\IEEEmembership{Fellow,~IEEE},
Rongrong Ji,~\IEEEmembership{Senior Member,~IEEE}
\thanks{Manuscript received Feb. XX, 2025.  (Corresponding author: Rongrong Ji)}
\thanks{This work was supported by National Science and Technology Major Project (No. 2022ZD0118202), the National Science Fund for Distinguished Young Scholars (No.62025603), the National Natural Science Foundation of China (No. U21B2037, No. U22B2051, No. U23A20383, No. 62176222, No. 62176223, No. 62176226, No. 62072386, No. 62072387, No. 62072389, No. 62002305 and No. 62272401), and the Natural Science Foundation of Fujian Province of China (No. 2021J06003, No.2022J06001).}
\thanks{W. Zhan and R. Ji are with the Key Laboratory of Multimedia Trusted Perception and Efficient Computing, Ministry of Education of China, Xiamen University, China. E-mail: zhanwy@stu.xmu.edu.cn, rrji@xmu.edu.cn}
  \thanks{M. Lin and S. Yan are with the Skywork AI, Singapore 118222. E-mail: linmb001@outlook.com, Shuicheng.yan@gmail.com}
}

\markboth{Journal of \LaTeX\ Class Files,~Vol.~14, No.~8, August~2021}%
{Shell \MakeLowercase{\textit{et al.}}: A Sample Article Using IEEEtran.cls for IEEE Journals}

\maketitle

\begin{abstract}
We introduce DiffusionTrend for virtual fashion try-on, which forgoes the need for retraining diffusion models. 
Using advanced diffusion models, DiffusionTrend harnesses latent information rich in prior information to capture the nuances of garment details. Throughout the diffusion denoising process, these details are seamlessly integrated into the model image generation, expertly directed by a precise garment mask crafted by a lightweight and compact CNN.
Although our DiffusionTrend model initially demonstrates suboptimal metric performance, our exploratory approach offers some important advantages:
(1) It circumvents resource-intensive retraining of diffusion models on large datasets.
(2) It eliminates the necessity for various complex and user-unfriendly model inputs.
(3) It delivers a visually compelling try-on experience, underscoring the potential of training-free diffusion model.
This initial foray into the application of untrained diffusion models in virtual try-on technology potentially paves the way for further exploration and refinement in this industrially and academically valuable field.
\end{abstract}


\section{Introduction}
\IEEEPARstart{V}{irtual} try-on technology~\cite{song2023image, vitonsurvey}, which digitally superimposes images of models wearing various outfits, represents a pivotal innovation in the fashion sector.
This advancement offers consumers an immersive and interactive experience with garments, allowing them to preview how clothing might look on them without physically visiting stores. 
For retailers, virtual try-on technology streamlines their operations by obviating the need to employ live models for merchandise display. It also circumvents the financial burdens associated with traditional product photography, leading to a marked improvement in operational efficiency. 
For commercial platforms, virtual try-on technology serves as a magnet for attracting a larger user base and fostering user loyalty. 
Collectively, virtual try-on is reshaping the landscape of the apparel retail industry, propelling the fashion sector toward a future characterized by heightened efficiency and customization.

\begin{table}[t]
\caption{Comparison of input requirements for previous virtual try-on models.
A checkmark (\Checkmark) indicates that the input modality is required, while a dash (-) indicates that it is not or not mentioned.}
\label{tab:intro}
\centering 
\renewcommand{\arraystretch}{1.2}

\resizebox{0.45\textwidth}{!}{
\begin{tabular}{c|ccccc}
\toprule
Method          & Clothes Mask & Densepose   & Segment Map & Clothes-Agnostic  & Keypoint\\
\midrule
TryOnDiffusion~\cite{zhu2023tryondiffusion}   
&  -           &  -           &  \Checkmark               & \Checkmark            & \Checkmark \\

DCI-VTON~\cite{gou2023dci} 
&    \Checkmark   &  \Checkmark      & \Checkmark              &  \Checkmark     &  - \\

LaDI-VTON~\cite{morelli2023ladi}   
&  \Checkmark            & \Checkmark    &   -              &  \Checkmark      &  - \\

WarpDiffusion~\cite{li2023warpdiffusion}  
& \Checkmark    &  -   &  -     
&   \Checkmark          & -  \\

OOTDiffusion~\cite{xu2024ootdiffusion} 
&    \Checkmark     &   -   & \Checkmark                & -         &\Checkmark    \\

StableVITON~\cite{kim2023stableviton} 
&   \Checkmark        &   \Checkmark         &  -               & \Checkmark            & -  \\

IDM-VTON~\cite{choi2024idm}    
&  \Checkmark       &   \Checkmark   &  -               &  \Checkmark           &  -  \\

Wear-Any-Way~\cite{chen2024wearanyway}   
&  \Checkmark       &  -   &  -   
&   \Checkmark          &  \Checkmark  \\

DiffusionTrend (Ours) & \Checkmark    & - & - & - & -
\\
\bottomrule
\end{tabular}}
\end{table}

Traditional virtual try-on solutions~\cite{choi2021vitonhd,lee2022high,ge2021parser,xie2023gpvton,fang2024pg,li2023povnet} are predicated on a two-stage pipeline utilizing Generative Adversarial Networks (GANs)~\cite{goodfellow2014gans}. The initial stage in this framework involves the application of an explicit warp module~\cite{zhang2024two,chou2021template,xu2021virtual,zhang2023limb,hu2022spg} to deform the clothing to the desired area on the body. The subsequent stage integrates the deformed clothing using a GAN-based try-on generator. 
To attain accurate clothing deformation, earlier methodologies~\cite{bai2022single, ge2021parser,han2019clothflow, lee2022high,xie2023gpvton, xing2022virtual,song2023fashion} have employed a trainable network designed to estimate a dense flow map~\cite{zhou2016view,du2022vton,yu2023vton}, thereby facilitating the mapping of the clothing onto the human form.
In parallel, various approaches~\cite{choi2021vitonhd,ge2021parser,xie2023gpvton,lee2022high,yang2020towards,issenhuth2020not, liu2021spatial,song2020unpaired}, have been proposed to address the misalignment issues between the warped clothing and the human body. Techniques such as normalization~\cite{choi2021vitonhd} and distillation~\cite{ge2021parser,issenhuth2020not} have been implemented to mitigate these discrepancies.
Recent advancements in diffusion models~\cite{ho2020ddpm} have led to a notable enhancement in the quality of image synthesis tasks, with a particular emphasis on the domain of virtual try-on.
In this context, contemporary research endeavors have leveraged pre-trained text-to-image diffusion models to produce high-fidelity results.
The TryOnDiffusion model~\cite{zhu2023tryondiffusion}, for instance, employs a dual U-Net architecture to perform the try-on task.
LADI-VTON~\cite{morelli2023ladi} and DCI-VTON~\cite{gou2023dci} either conceptualize clothing items as pseudo-words or integrate garments through the use of warping networks into pre-trained diffusion models.

\begin{figure*}[t]
    \centering
    \includegraphics[width=0.98\linewidth ]{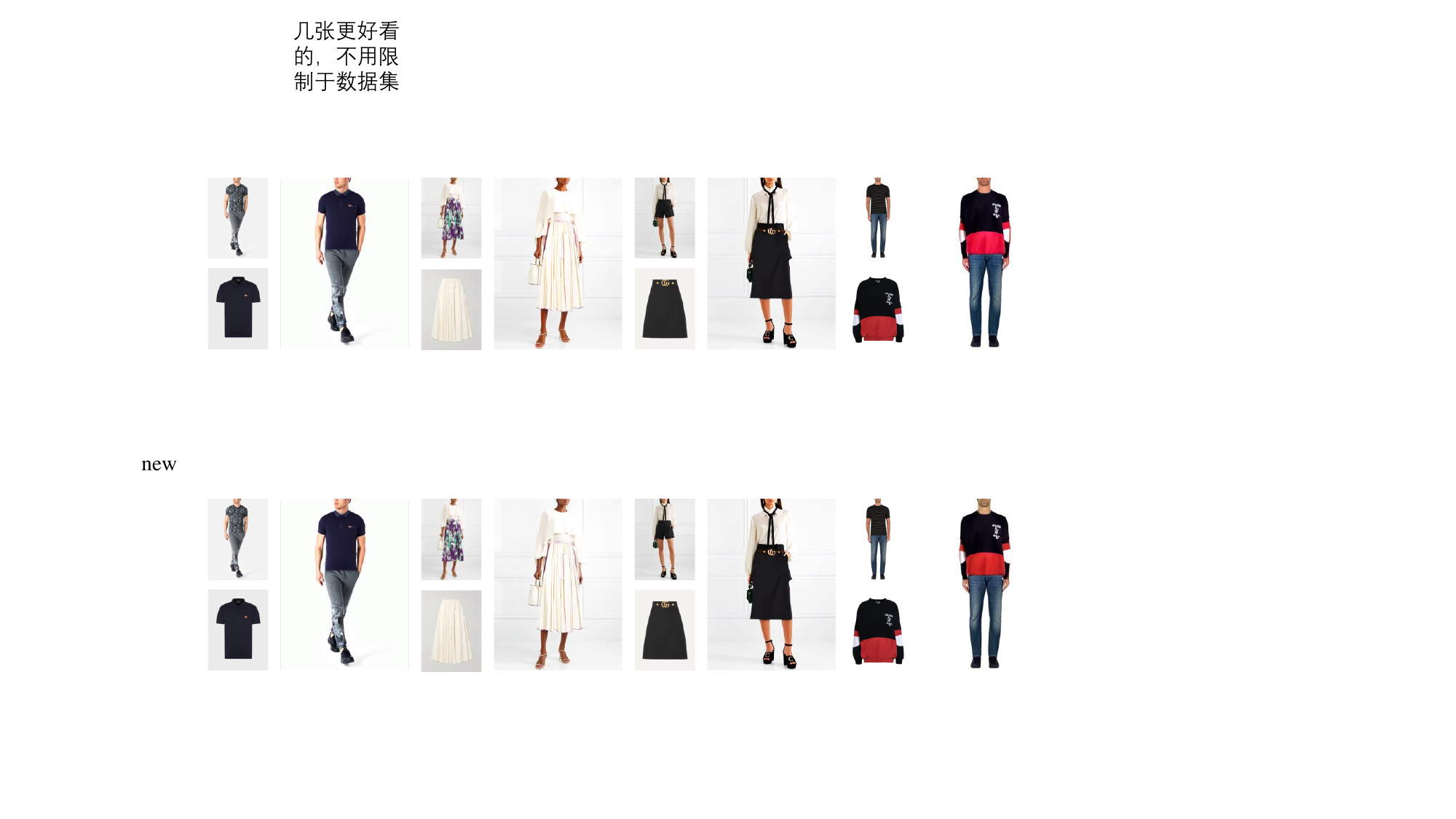}
    \caption{
    Images generated by the proposed DiffusionTrend model, given an input target model and a try-on clothing item both from DressCode dataset~\cite{morelli2022dresscode}.}
   \label{fig:first}
\end{figure*}

Despite these advancements, a review of existing studies reveals that the high precision and realism achieved by current methods require training on extensive try-on datasets~\cite{morelli2022dresscode}, particularly in diffusion-based approaches~\cite{zhu2023tryondiffusion, morelli2023ladi, gou2023dci}. 
Diffusion models simulate the data generation process by iteratively modeling the diffusion and reverse diffusion steps. Each iteration involves complex probability distribution calculations and requires substantial computational resources, leading to considerable overall training costs.
In the words, current methods rely on resource-intensive network training.
While intensive training is well-suited for generating complex poses and capturing fine image details, the substantial computational costs can make it an unattractive option.
Besides, as shown in Table\,\ref{tab:intro}, these methods often require multiple auxiliary modules or APIs as model inputs, such as densepose~\cite{guler2018densepose}, segmentation maps~\cite{gong2017humanparser}, clothes-agnostic images/masks/representations~\cite{han2018viton} and keypoints~\cite{cao2017openpose}.
The integration of these tools not only increases inference overhead but also suffers from insufficient encapsulation in some cases, leaving non-professional users to deal with complex configurations and unstandardized interfaces, which significantly hampers accessibility and user experience.
There is a pressing need for the community to explore more accessible and less resource-intensive methods, with the effort of less compromise on the quality of image generation---a domain that, to our knowledge, remains uncharted to date.

In marked contrast, our proposed DiffusionTrend offers a streamlined, lightweight training approach that circumvents the need to retrain a diffusion model, thus liberating it from the reliance on expensive and resource-intensive computational infrastructure. 
This technique reduces inference overhead and simplifies the workflow by dispensing with the requirement for intricate segmentation, pose extraction, and other preliminary processing steps for the input images, leading to a more accessible and economical solution.
Specifically, a lightweight, compact CNN is utilized to outline the clothing in both the model and garment images. 
It harmonizes image and textual features and conducts clustering at the feature level to produce effective masks. 
Subsequently, we make full use of the latents derived from DDIM inversion~\cite{song2020ddim}, which are replete with prior information and act as superior conduits for the detailed features of the garment. 
During the early stages of the diffusion denoising process, the seamless integration of the target garment into the model's image reconstruction is achieved by blending the latent representations of both the model and the garment.
In the subsequent stages of denoising, leveraging the self-repairing capabilities of the pre-trained latent diffusion model, we maintain the model's identity and background coherence by selectively replacing the latents in the background areas. This approach ensures that the garment is harmoniously merged into the overall image.
To further enhance robustness under diverse generation conditions, we apply an adaptive sampling termination strategy based on perceptual color difference to automatically determine the optimal number of sampling steps during inference.
Fig.\,\ref{fig:first} shows several examples of our proposed DiffusionTrend.

Despite its suboptimal metric performance, our initial exploration with DiffusionTrend has yielded several valuable contributions that could indeed pave the way for further research and refinement in the realm of untrained diffusion models for virtual try-on technology:
\begin{itemize}
    \item It eliminates the necessity for resource-intensive training of diffusion models on extensive datasets, thereby reducing the computational demands.
    \item It removes the need for cumbersome and user-unfriendly model inputs, reducing computational costs and streaming the pipeline of virtual try-on.
    \item It delivers a visually appealing virtual try-on experience, highlighting the potential of diffusion models that do not require training for future research. 
\end{itemize}

These contributions underscore the significance of our DiffusionTrend model as a foundation for future advancements in virtual try-on technology, emphasizing the potential of exploring a training-free approach.

\section{Related Work}
\label{related_work}
\subsection{Image Editing through Diffusion Processes}
The practice of integrating specific content into a base image to produce realistic composites is prevalent in image editing leveraging diffusion processes.
Initially, the field was dominated by text-based models for image editing~\cite{brooks2023instructpix2pix,kawar2023imagic}.
InstructPix2Pix~\cite{brooks2023instructpix2pix} employs paired data to train diffusion models capable of generating an edited image from an input image and a textual instruction.
Conversely, Imagic~\cite{kawar2023imagic} harnesses a pre-trained text-to-image diffusion model to generate text embeddings that align with both the input image and the target text.
The abstract nature of text poses a limitation in accurately delineating the subtleties of objects, therefore, image conditioning was introduced to offer more concrete and precise descriptions.
DCFF~\cite{xue2022dccf} pioneers the use of pyramid filters for image composition, which was subsequently advanced by Paint by Example~\cite{yang2023paintbyexample}, employing CLIP embeddings of the reference image to condition the diffusion model.
Most contemporary methodologies, such as Dreambooth~\cite{ruiz2023dreambooth} (all model parameters), Textual Inversion~\cite{gal2022textualinversion} (a word vector for novel concepts), and Custom-Diffusion~\cite{kumari2023custom} (cross-attention parameters), rely heavily on fine-tuning techniques.
In contrast, a handful of approaches~\cite{hertz2022p2p,cao2023masactrl} adopts a training-free paradigm.
Prompt-to-prompt~\cite{hertz2022p2p} modifies the input text prompt to steer the cross-attention mechanism for nuanced image editing, while MasaCtrl~\cite{cao2023masactrl} employs a mask-guided mutual attention strategy for non-rigid image synthesis and editing.
These training-free methods offer a cost-effective alternative, eliminating extensive training while still delivering commendable generative outcomes.

\subsection{Virtual Try-on with Diffusion Models}

Diffusion models have demonstrated remarkable efficacy in the domain of image editing, with image-based virtual try-on representing a specialized subset of these tasks, contingent upon a specific garment image.
Adapting text-to-image diffusion models to accommodate images as conditions, is straightforward, but the spatial discrepancies between the garment and the subject's pose challenge the fidelity of texture details in the virtual try-on outcomes~\cite{li2023warpdiffusion,gou2023dci,morelli2023ladi}.
Methodologies such as WarpDiffusion~\cite{li2023warpdiffusion}, DCI-VTON~\cite{gou2023dci}, and LADI-VTON~\cite{morelli2023ladi} conceptualize clothing as pseudo-words, employing warping techniques through CNN networks to adjust clothing to various poses, yielding satisfactory results.
TryOnDiffusion~\cite{zhu2023tryondiffusion} employs a dual U-Net architecture for the virtual try-on task, implicitly conducting garment warping through the interplay between cross-attention layers. This approach effectively resolves the issue of texture misalignment without the need for a dedicated warp module. 
Similarly, the StableVITON~\cite{kim2023stableviton} incorporates zero cross-attention blocks to condition the intermediate feature maps of a spatial encoder, thereby circumventing the requirement for a warp module. 
The Wear-Any-Way~\cite{chen2024wearanyway} enhances the process of virtual garment alteration, providing more adaptable control over the manner in which clothing is depicted.
The IDM-VTON~\cite{choi2024idm} enhances the virtual try-on process by integrating attention mechanisms and high-level semantic encoding into the diffusion model.

\section{Methodology}
\label{method}

\subsection{Preliminaries}

\textbf{Latent Diffusion Models.}
Latent diffusion models (LDMs) use an encoder $\mathcal{E}$ to convert an input image $x_0 \in \mathbb{R}^{H \times W \times 3}$ into a lower-dimensional $z_0 = \mathcal{E}(x_0) \in \mathbb{R}^{h \times w \times c}$. 
Here, the downsampling ratio is $f = H/h = W/w$, and $c$ is the channel number. The forward diffusion process is:
\begin{equation}
    z_t = \sqrt{\bar{\alpha}_t} z_0 + \sqrt{1-\overline{\alpha}_t} \epsilon , \quad \epsilon \sim \mathcal{N}(0,I),
\end{equation}
where $\{\alpha_t\}_{t=1}^T$ denotes variance schedules, with $\bar{\alpha}_t = \prod_{i=1}^t \alpha_i$. A U-Net $\epsilon_\theta$ refines noise estimation. This is crucial for reconstructing the latent representation $z_0$ from the initial noisy state $z_T \sim \mathcal{N}(0, \mathbf{I})$:
\begin{equation}
\label{equ:denoise}
    z_{t-1} = \sqrt{\frac{\alpha_{t-1}}{\alpha_t}} z_t + \Bigg(\sqrt{\frac{1}{\alpha_{t-1}}-1} - \sqrt{\frac{1}{\alpha_t}-1} \Bigg) \cdot \epsilon_\theta \big(z_t,t,\tau_\theta (\mathcal{P}) 
    \big).
\end{equation}

The text encoder $\tau_\theta(\mathcal{P})$ converts text prompt $\mathcal{P}$ into an embedding that is integrated with the U-Net's intermediate noise representation using cross-attention mechanisms. At time step $t = 0$, the decoder $\mathcal{D}$ transforms the latent space representation $z_0$ back into the original image domain $x_0 = \mathcal{D}(z_0)$.

\textbf{DDIM Inversion.}
DDIM inversion uses DDIM sampling~\cite{song2020ddim} to ensure deterministic sampling by setting the variance in Eq.\,(\ref{equ:denoise}). 
It assumes the reversibility of the ordinary differential equation~\cite{chen2018neuralode} through incremental steps, ensuring a controlled transition from initial state $z_0$ to final noise $z_T$:
\begin{align}\label{inversion}
z^{*}_t = &\sqrt{\frac{\alpha_t}{\alpha_{t-1}}} z^*_{t-1} -\sqrt{\frac{\alpha_t}{\alpha_{t-1}}}  \Bigg(\sqrt{\frac{1}{\alpha_{t-1}}-1} - \sqrt{\frac{1}{\alpha_t}-1}\Bigg) \nonumber \\
   & \cdot \epsilon_\theta \big(z^*_{t-1},t-1,\tau_\theta (\mathcal{P}) \big) ,
\end{align}

We start with the noisy latent state $z^*_T$ and proceed with denoising as outlined in Eq.\,(\ref{equ:denoise}). This method approximates $z^*_0$, which closely resembles the original latent representation $z_0$.
Our goal is to incorporate information from a garment image $I^g$ into the reconstruction of the model image $I^m$, depicting the model wearing the garment from $I^g$.

To address the challenge of extensive training overhead and various model inputs, we choose not to alter any weights or structures of the pre-trained diffusion model. Instead, we introduce a lite-training visual try-on method called ``DiffusionTrend''.
Our methodology consists of three stages. 
1) A lightweight and compact CNN accurately delineates the apparel in both the model and garment images.
2) At an appropriate point in the process, garment details are integrated into the reconstruction phase of the model image.
3) To ensure the coherence of the generated background with the model, we use a latent substitution technique to restore the background, leveraging the diffusion model's restorative properties to blend it seamlessly with the newly rendered apparel.
A comprehensive discussion of the first stage is in Sec.\,\ref{sec:mask}, while the latter two stages are discussed in Sec.\,\ref{sec:reconstruction}.

\subsection{Precise Apparel Localization}
\label{sec:mask}

In conventional GAN-based~\cite{lee2022high, ge2021parser, xie2023gpvton} or current Diffusion-based~\cite{zhu2023tryondiffusion,gou2023dci,morelli2023ladi} try-on methods, a precise clothing mask is crucial. 
This mask ensures correct apparel placement on the model and accurate extraction of garment features while avoiding interference from non-garment regions. 

\begin{figure}[!b]
    \centering
    \includegraphics[width=0.5\textwidth]{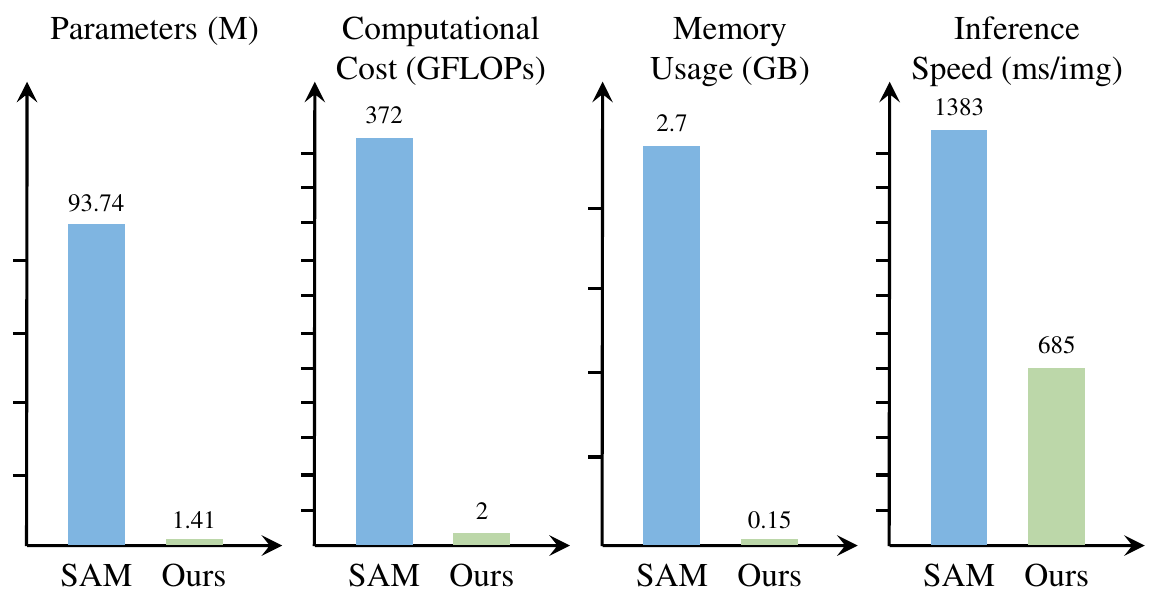}
    \caption{
    Inference cost comparison: Segment Anything Model (SAM)~\cite{kirillov2023segmentanything} \emph{vs}. our Apparel Localization Network.}
    \label{fig:cost}
\end{figure}

\begin{figure*}[!t]
    \centering
    \includegraphics[width=\textwidth]{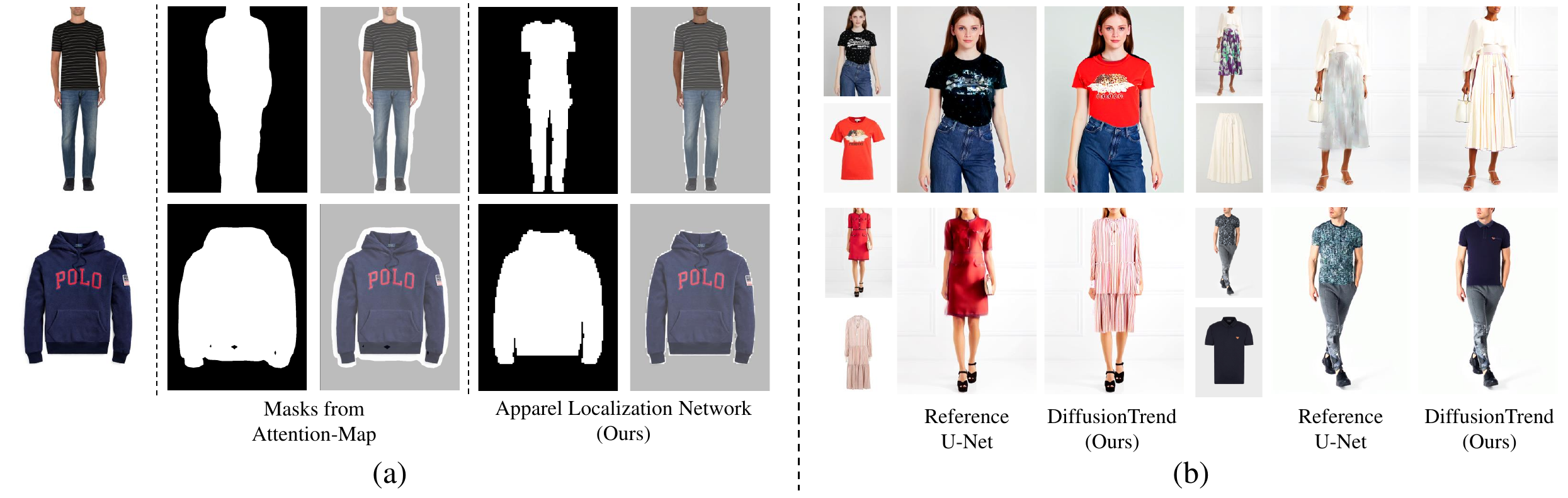}
    \caption{
    {Visual comparison of mask generation and try-on results. (a) Precision of apparel localization masks generated by our network compared to masks from attention maps. (b) Quality comparison of generation results: Reference U-Net method \emph{vs}. DiffusionTrend.
    }}
    \label{fig:attn_ref}
\end{figure*}

Traditional segmentation models~\cite{he2017maskrcnn,kirillov2023segmentanything} excel in mask generation tasks. However, after thorough evaluation, we have chosen not to employ pre-built models like Segment Anything Model  (SAM)~\cite{kirillov2023segmentanything}. Instead, we propose a lightweight CNN tailored for our application. This decision is informed by the following considerations:

\textbf{Resource Efficiency in Constrained Environment:}
While models like SAM offer outstanding performance, they come with a significant computational overhead, particularly in terms of memory usage and processing time. As shown in Fig.\,\ref{fig:cost}, SAM requires 2.7 GB of GPU memory and 372 GFLOPs to process an image of size 768$\times$1024. In contrast our lightweight CNN requires a mere 0.15 GB of memory and 2 GFLOPs, making a substantial reduction in resource usage. In practical virtual try-on applications, especially in online multi-user settings, servers are tasked with managing a high volume of concurrent requests. In such scenarios, efficient resource use is more critical than slight gains in segmentation precision. Our lightweight CNN not only minimizes operational expenses but also significantly improves response time.

\textbf{Task-Specific Optimization Over General-Purpose Segmentation:}
Off-the-shelf models like SAM are engineered for a wide range of semantic segmentation tasks and lack customization for the nuances of virtual try-on applications. Our proposed lightweight CNN, on the other hand, incorporates task-specific textual cues (\emph{e.g.}, ``top", ``skirt") to enable precise segmentation of clothing areas, sidestepping irrelevant regions.
We do not adopt the CLIP~\cite{radford2021clip} image encoder, as it is primarily optimized for global image-text alignment and lacks the spatial resolution required for pixel-accurate apparel segmentation. Moreover, it introduces additional inference cost, which contradicts our lightweight design principle.

\textbf{Balancing Computational Load Across the Pipeline:}
The diffusion model at the core of our system is inherently resource-intensive. Incorporating a high-cost segmentation model like SAM would further strain computational resources, contracting our aim of creating a lightweight and cost-effective virtual try-on solution. Our lightweight CNN not only harmonizes with diffusion, but also supports our overarching goal of developing an economical segmentation network for this specific domain. Using a generic segmentation model would compromise this objective and diminish the distinctive value of our research.

In addition, we have attempted to extract masks using the intermediate attention map of a diffusion U-Net. During the inversion process, we utilize the concept of ``clothes'' to interact with image features via cross-attention. However, the extracted masks, we find, fail to meet the precision requirements for virtual try-on tasks. As shown in Fig.\,\ref{fig:attn_ref}(a), the attention map roughly identifies the garment region but often includes extraneous parts.

\begin{figure*}[t]
    \centering
    \includegraphics[width=0.95\textwidth]{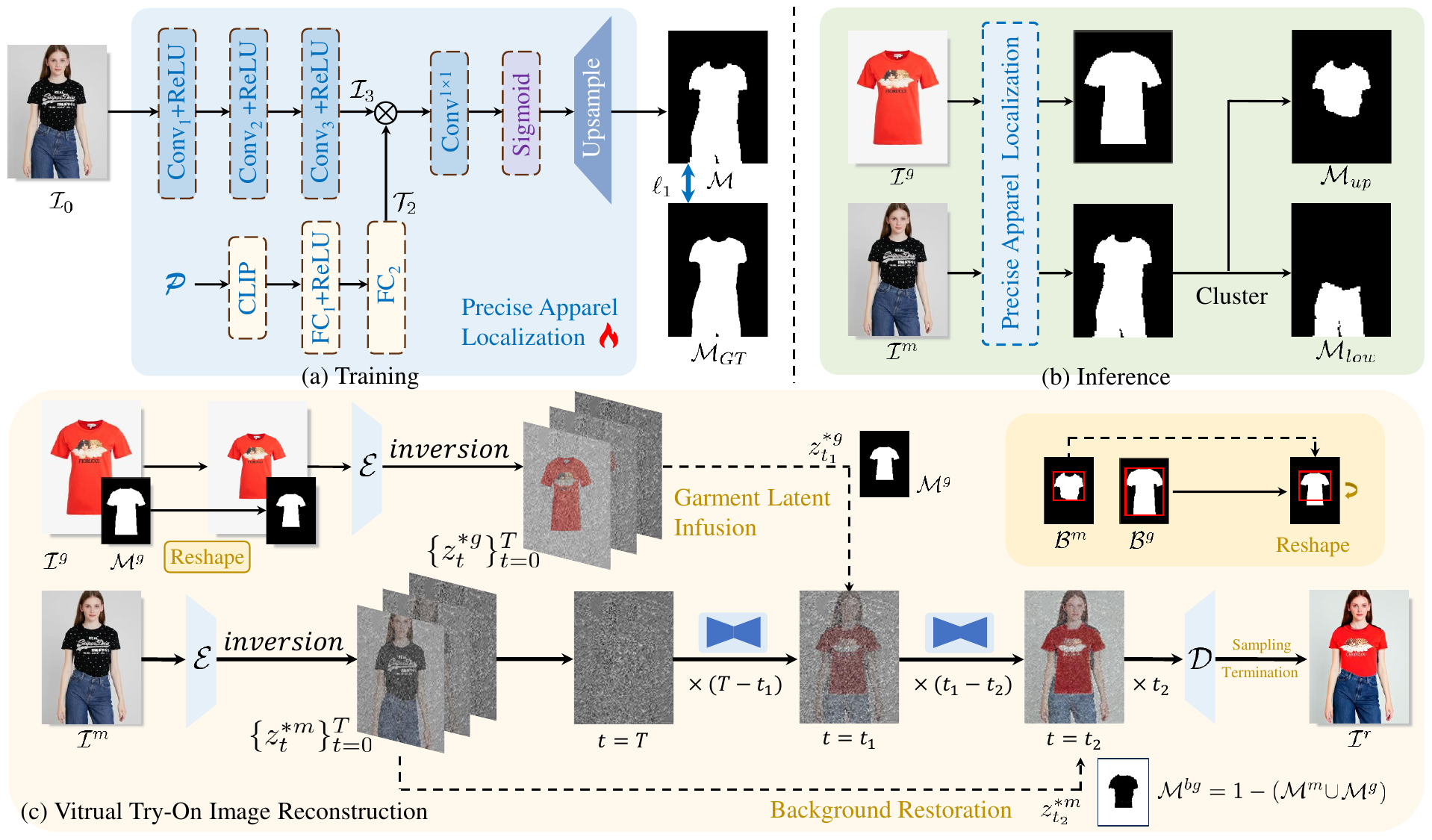}
    \caption{DiffusionTrend framework. 
    (a) A lightweight apparel localization network to predict precise garment masks. 
    (b) The network inference. 
    (c) The reconstruction of the virtual try-on image.}
    \label{fig:pipeline}
\end{figure*}

Therefore, in the consideration of balancing accuracy and saving computational resources, we have engineered a compact CNN in Fig.\,\ref{fig:pipeline}(a) for precise apparel localization.
By combining textual and visual features, we minimize manual intervention. 
Users only need to specify the target category (\emph{e.g.}, upper garments, lower garments, or dresses), and the model automatically generates precise mask outputs.
Our network accepts a model image $\mathcal{I}_0$ as input and processes it through three $3 \times 3$ $Conv^{3 \times 3}_i$ layers with $ReLU$ activation, culminating in the image features $\mathcal{I}_3$, as:
\begin{equation}
     \mathcal{I}_{i} = ReLU\big({Conv}^{3 \times 3}_i(\mathcal{I}_{i-1})\big), \quad i = 1, 2, 3.
\end{equation}

An apparel-related prompt $\mathcal{P}$ is transformed into a text embedding $\mathcal{T}_{0} = Clip(\mathcal{P})$ by a $Clip$~\cite{radford2021clip} text encoder with fixed parameters. 
The text embedding $\mathcal{T}_{0}$ is then advanced through two $FC_i$ layers with a $ReLU$ function in between, to produce the text-derived features $\mathcal{T}_2$ as:
\begin{equation}
     \mathcal{T}_2 = FC_2\Big(ReLU\big(FC_1(\mathcal{T}_0)\big)\Big).
\end{equation}

Next, we amalgamate the image feature $\mathcal{I}_3$ with the text feature $\mathcal{T}_2$ using a $1 \times 1$ $Conv^{1 \times 1}$ layer, followed by $Sigmoid$ and $Upsample$ functions as:
\begin{equation}
    \mathcal{M} = Upsample \Big( Sigmoid \big(Conv^{1 \times 1}(\mathcal{I}_3 \cdot \mathcal{T}_2 )\big)\Big).
\end{equation}

For the training, we compute the $\ell_1$ norm between $\mathcal{M}$ and ground-truth mask $\mathcal{M}_{GT}$, employing this as the loss function to refine the network's parameters. Note that the predicted $\mathcal{M}$ serves as a mask for all clothing items on the model, which may encompass both upper and lower garments.

To delineate between the upper and lower garments and extract their respective masks, we perform K-means clustering upon the masked features $\mathcal{M} \cdot \mathcal{I}_3$. During the clustering process, the cluster with the smallest mean feature value is typically associated with the background and is excluded from further consideration. The remaining clusters are then analyzed based on the vertical positions of their centroids, enabling effective separation of top and bottom garment masks:
\begin{equation}
    \mathcal{M}_{up}, \,\, \mathcal{M}_{low} = Cluster(\mathcal{M} \cdot \mathcal{I}_3).
\end{equation}

It should be noted that for models adorned in addresses or when processing a garment image, $\mathcal{M}$ is utilized directly as the mask, thereby obviating the need for clustering.

To further enhance the precision and boundary regularity of the K-means-generated masks, we apply a classic active contour-based~\cite{Kass88} refinement method as a post-processing step. This technique iteratively optimizes the mask boundary by minimizing an energy function that balances edge attraction and contour smoothness, producing more natural and connected garment regions. A comparative study of garment mask refinement strategies is presented in Sec.\,\ref{sec:ablation}.

Our apparel localization network is notably lightweight, making the training process highly cost-effective when compared to methods~\cite{li2023warpdiffusion, kim2023stableviton, gou2023dci} that require training a diffusion model. It takes us only 20 hours to process the entire DressCode dataset~\cite{morelli2022dresscode} using two RTX 3090 GPUs.
As shown in Fig.\,\ref{fig:cost}, our proposed localization network operates fully automatically with only 2.00 GFLOPs and 0.15G of memory, and its accuracy analysis can be found in Sec.\,\ref{sec:ablation}.

\subsection{Virtual Try-On Image Reconstruction}
\label{sec:reconstruction}

Herein, we delve into the process of integrating garment details into the reconstruction phase of the model image.
As illustrated in Fig.\,\ref{fig:pipeline}(c), our virtual try-on image reconstruction process encompasses two critical stages: the infusion of garment latents and the restoration of the background.

\textbf{Garment Latent Infusion.}
Recent studies~\cite{choi2024idm,xu2024ootdiffusion} indicate that employing U-Net for feature extraction necessitates treating it as a high-parameter module, which entails considerable training expenses. This approach conflicts with our research goal of achieving a training-free diffusion model solution, as such high computational costs undermine the feasibility of lightweight virtual try-on applications.
We first experiment with cross-attention-based MasaCtrl~\cite{cao2023masactrl}, which involves a reference U-Net extracting garment features and performing key-value exchanges with the main U-Net during the attention stage. 
Although this method avoids additional training, it heavily relies on dual-branch U-Nets and attempts to guide attention interactions with masks derived from our apparel localization network. 
As shown in Fig.\,\ref{fig:attn_ref}(b), the results fall short in accurately rendering fine-grained garment details. We surmise that direct feature injection through cross-attention, while effective at generating semantically coherent content, lacks the ability to capture the intricate details essential for high-quality virtual try-on results.

Given these challenges, we have been iteratively refining our methodology. After extensive exploration, we develop the approach illustrated in Fig.\,\ref{fig:pipeline}(c), which achieves a precise and efficient fusion of garment details with model images while maintaining the lightweight and training-free objectives.

Given a model image $\mathcal{I}^m$ and a garment image $\mathcal{I}^g$, we utilize our apparel localization network to acquire the masks $\mathcal{M}^m$ and $\mathcal{M}^g$.
A natural thought is to leverage an explicit warp module to align the in-store garment $\mathcal{I}^g$ with the pose, position, and content of the clothing in $\mathcal{I}^m$.
Initially, we adopted three representative methods, including the warping modules from DCI-VTON~\cite{gou2023dci}, SCW-VTON~\cite{han2024scwvton}, and PL-VTON~\cite{zhang2023plvton}, to preprocess garment images.
However, the results were unsatisfactory. In Fig.\,\ref{fig:warping}, the warped garment images introduce severe distortions into the inversion and reconstruction phases. This, in turn, leads to highly distorted virtual try-on results, significantly degrading the overall quality of the pipeline.

We attribute this to incompatibility between the independent warp module and the untrained U-Net in our training-free paradigm, with detailed analysis provided in Sec.\,\ref{sec:ablation}.
Consequently, we opt for perspective transformations to address the alignment problem. By computing the bounding boxes $\mathcal{B}^m$ and $\mathcal{B}^g$ for $\mathcal{M}^m$ and $\mathcal{M}^g$, respectively, we reshape the in-store garment image $\mathcal{I}^g$ to align with the position and size of the clothing in the model image $\mathcal{I}^m$.

\begin{figure*}[!t]
    \centering
    \includegraphics[width=\textwidth]{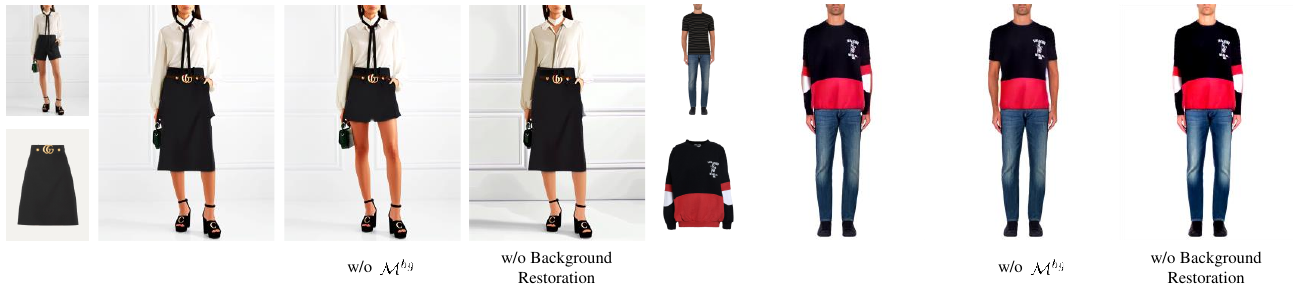}
    \caption{
Ablation studies on $\mathcal{M}^{bg}$ and background restoration are presented. Omitting $\mathcal{M}^{bg}$ results in inaccuracies such as the reconstruction of long skirts as short skirts and long sleeves as short sleeves.
Without background restoration, various issues arise, including altered background colors, distorted facial features, and unintended changes to other parts of the model's attire.
}
    \label{fig:sup1}
\end{figure*}

The reshaping process starts by resizing the bounding box $\mathcal{B}^g$ to match $\mathcal{B}^m$, achieving initial positional alignment.
However, this direct resizing may lead to distortions in the garment's proportions, affecting its style and fit (\emph{e.g.}, changes in pant length or skirt length). To address this, we analyze the inherent features of the original garment, such as shoulder width, waistband, or hemline position. These features help estimate appropriate scaling factors for different garment types, ensuring that key attributes (\emph{e.g.}, shirt length, pant length, or dress silhouette) are preserved. Using these scaling factors, we apply proportional stretching adjustments to restore the garment's original design and fit.

Following the reshaping step, we apply appropriate perspective transformations to simulate simple rotational and deformational effects. For instance, the left side of the garment can be rotated inward to mimic the perspective of a model turning to the left. These transformations simulate basic pose variations, addressing alignment and deformation challenges.

Once aligned, the model image $\mathcal{I}^m$ and the adjusted garment image $\mathcal{I}^g$ are transformed into their latent representations $z_0^m = \mathcal{E}(\mathcal{I}^m)$ and $z_0^g = \mathcal{E}(\mathcal{I}^g)$, 
where $\mathcal{E}$ denotes the encoder. These latents are then subjected to the DDIM Inversion process in Eq.\,(\ref{inversion}), yielding noisy latent sets $\{z^{*m}_t\}_{t=0}^T$ and $\{z^{*g}_t\}_{t=0}^T$.

Starting from $z^{*m}_T$, we reconstruct the model image using Eq.\,(\ref{equ:denoise}) where the latent at the $t$-th time step is $z_t^m$. Our goal is to decode an image of the model wearing the garment from image $\mathcal{I}^g$, thereby achieving a harmonious fusion of the model’s appearance and the desired attire.
A literature review~\cite{wu2023freeinit} has shown that latents derived from the inversion process retains rich prior information. This property makes it ideal for capturing the intricate features of the target garment.
An effective strategy is to integrate the garment latent $z_{t_1}^{*g}$, preserved during the inversion stage, into the masked region $M^g$ at time step $t_1$. This occurs early in the denoising process as:
\begin{equation}
    z^*_{t_1} = z^m_{t_1} \cdot (1 - \mathcal{M}^g) + z^{*g}_{t_1} \cdot \mathcal{M}^g.
\end{equation}

The denoising continues with the infused latent. It ensures the seamless information transferred from the garment image to the model image, and achieves a precise attire on the model.
We choose latent-space fusion over image-space masking, as it enables progressive integration of garment details and avoids visual artifacts caused by hard region cuts.

While perspective transformations cannot simulate the complex deformations in real try-on scenarios, they offer a simple and lightweight solution for basic alignment, which aligns with our goal of a training-free virtual try-on framework and lays the groundwork for future advancements in garment alignment.

\textbf{Background Restoration.}
While the garment latent infusion yields significant results, it negatively impacts the generation of background content in subsequent denoising steps.
In Fig.\,\ref{fig:sup1}, issues such as altered background color, distorted facial features, and unintended changes to other parts of the model's attire occur.
We speculate that this may be due to the prompt ``a model wearing clothes'' used during the generation process, which guides the reconstruction to focus on maintaining the garment's structure, while the background, lacking specific guidance, becomes more susceptible to distortion.
To address this, we must implement strategies to restore the background and preserve the model's identity and background information.

Motivated by this, we inject the model latent $z^m_{t_2}$ into the regions outside the model clothing mask $\mathcal{M}^m$ at time step $t_2$, a later stage in the diffusion denoising process focused on generating detailed information. 
Leveraging the diffusion model's inherent repair capability, a few subsequent denoising steps integrate the background latents with the target garment. This process can be formalized as:
\begin{equation}
\label{equ:mask_bg_1}
    z^*_{t_2} = z_{t_2}^m \cdot \mathcal{M}^m  + z^{*m}_{t_2} \cdot (1 - \mathcal{M}^m).
\end{equation}

After extensive experiments, we found that using $\mathcal{M}^m$ to differentiate between the foreground and background is not optimal. 
As shown in Fig.\,\ref{fig:sup1}, if the original model is wearing a short-sleeved garment and the target garment is long-sleeved, the sleeve in the generated image is incorrectly marked as background and replaced with the arm from the original model image $I^m$, causing a style mismatch.
To solve this, we use the union of the model clothing mask $\mathcal{M}^m$ and the garment mask $\mathcal{M}^g$, and the complement as the background mask $\mathcal{M}^{bg}$:
\begin{equation}
    \mathcal{M}^{bg} = 1- (\mathcal{M}^m \cup \mathcal{M}^g).
\end{equation}

Consequently, Eq.\,(\ref{equ:mask_bg_1}) is revised to the following:
\begin{equation}
z^{*}_{t_2} = z^{m}_{t_2} \cdot (1 - \mathcal{M}^{bg}) + z^{*m}_{t_2} \cdot \mathcal{M}^{bg}.
\end{equation}

The subsequent denoising steps proceed on $z^{*}_{t_2}$ until the latent $z_0$ is reached according to Eq.\,(\ref{equ:denoise}). Ultimately, by decoding $z_0$, we obtain the generated try-on result image $\mathcal{I}^r$.

\begin{table*}[!t]
\setlength{\tabcolsep}{5pt}
\caption{Quantitative results on the VITON-HD and DressCode datasets.}
\label{tab:result}
\centering
\renewcommand{\arraystretch}{1.3}
\setlength{\tabcolsep}{4mm}{
\resizebox{\linewidth}{!}{
\begin{tabular}{c|cccc|cccc}
\toprule
\multirow{2}{*}{Method} & \multicolumn{4}{c|}{VITON-HD} & \multicolumn{4}{c}{DressCode} \\
\cline{2-9}
                        & LPIPS$\downarrow$ & SSIM$\uparrow$ & FID$\downarrow$ & KID$\downarrow$ & LPIPS$\downarrow$ & SSIM$\uparrow$ & FID$\downarrow$ & KID$\downarrow$ \\
\toprule
TryOnDiffusion~\cite{zhu2023tryondiffusion}          & -                 & -              & 13.447           & 6.964           & -                 & -              & -               & -               \\
DCI-VTON~\cite{gou2023dci}                & 0.0530            & 0.8920         & 9.130            & 0.870           & 0.0443            & -              & 11.800          & -               \\
LaDI-VTON~\cite{morelli2023ladi}               & 0.0910            & 0.8760         & 9.410            & 0.160           & 0.0640            & 0.9060         & 6.480           & 0.220           \\
WarpDiffusion~\cite{li2023warpdiffusion}           & 0.0880            & 0.9850         & 8.610            & -               & 0.0890            & 0.9010         & 9.187           & -               \\
OOTDiffusion~\cite{xu2024ootdiffusion}            & 0.0710            & 0.8780         & 8.810            & 0.820           & 0.0450            & 0.9270         & 4.200           & 0.370           \\
StableVITON~\cite{kim2023stableviton}             & 0.0732            & 0.8880         & 8.233            & 0.490           & 0.0388            & 0.9370         & 9.940           & 0.120           \\
IDM-VTON~\cite{choi2024idm}                & 0.1020            & 0.8700         & 6.290            & -               & 0.0620            & 0.9200         & 8.640           & -               \\
Wear-Any-Way~\cite{chen2024wearanyway}            & 0.0780            & 0.8770         & 8.155            & 0.780           & 0.0409            & 0.9340         & 11.720          & 0.330           \\

DiffusionTrend (Ours)    & 0.0918            & 0.8592         & 10.433           & 0.540           & 0.0720            & 0.9172         & 9.704          & 0.431         \\
\toprule
\end{tabular}}
}

\end{table*}

\subsection{Adaptive Sampling Termination}
\label{sec:termination}

While DiffusionTrend achieves training-free virtual try-on by injecting garment and background latents at predefined timesteps, we observe that the quality of the generated results is sensitive to the total number of sampling steps $T$, especially in terms of color consistency and structural coherence. To mitigate this issue and enhance robustness across diverse generation conditions, we introduce a sampling termination strategy based on the perceptual color difference metric, CIEDE2000~\cite{sharma2005ciede2000} ($\Delta E_{00}$). This strategy allows the model to dynamically determine the optimal termination step during sampling.

Specifically, we compute the average $\Delta E_{00}$ between the generated image and the target garment image in the garment-masked region after each sampling step. Once the color deviation increases relative to the previous step, it indicates that the generative result has started to deviate from the target appearance. At this point, we halt the sampling process and roll back to the previous intermediate result with the lowest $\Delta E_{00}$ value. This approach requires no additional training or parameter tuning and effectively balances quality and efficiency.

\begin{figure*}[p]
    \centering
    \includegraphics[width=0.95\textwidth]{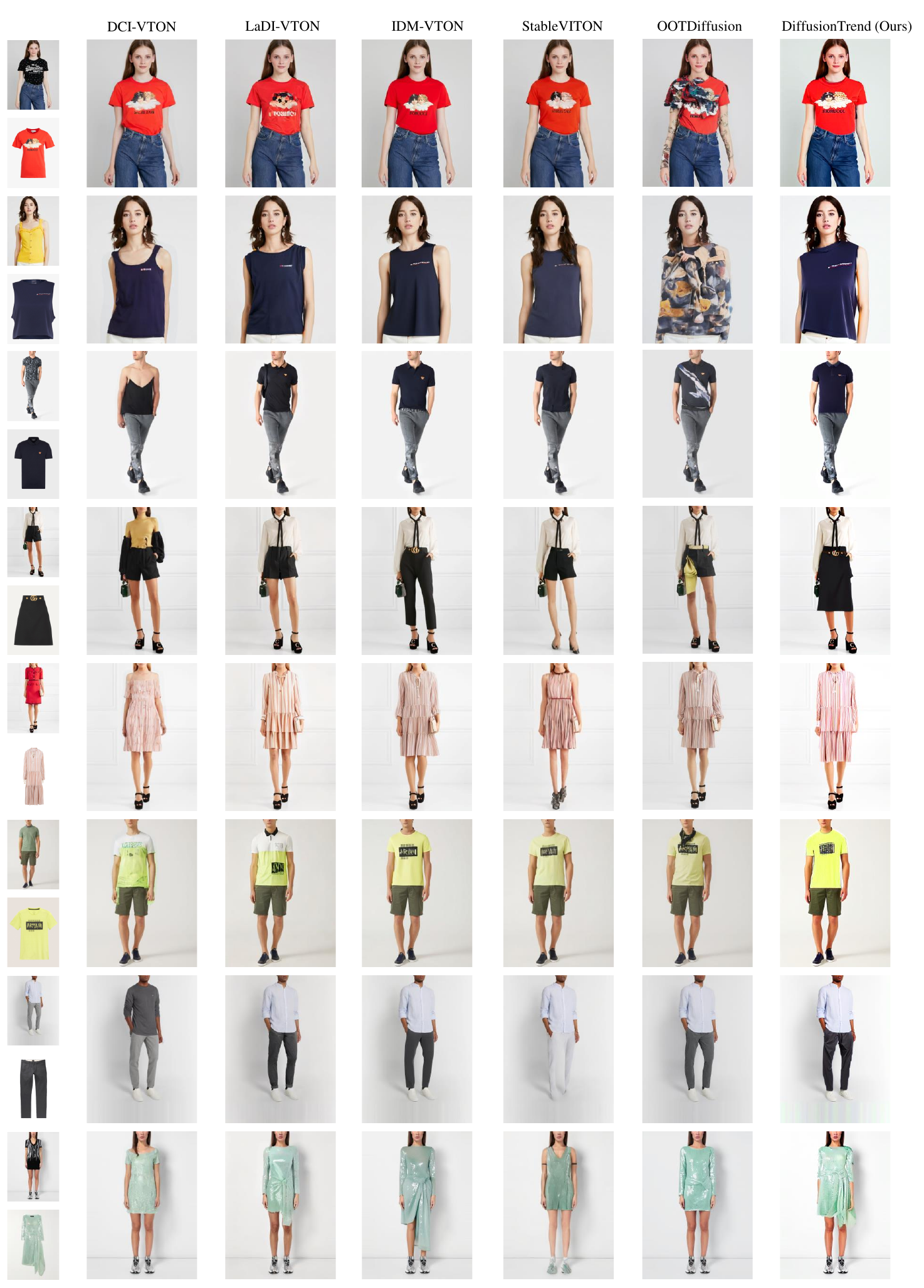}
    \caption{Qualitative comparisons on VITON-HD~\cite{choi2021vitonhd}(1st $\sim$ 2nd row) and DressCode~\cite{morelli2022dresscode} (3rd $\sim$ 8th row).
    }
    \label{fig:show1}
\end{figure*}

\begin{figure*}[t]
    \centering
    \includegraphics[width=0.85\textwidth]{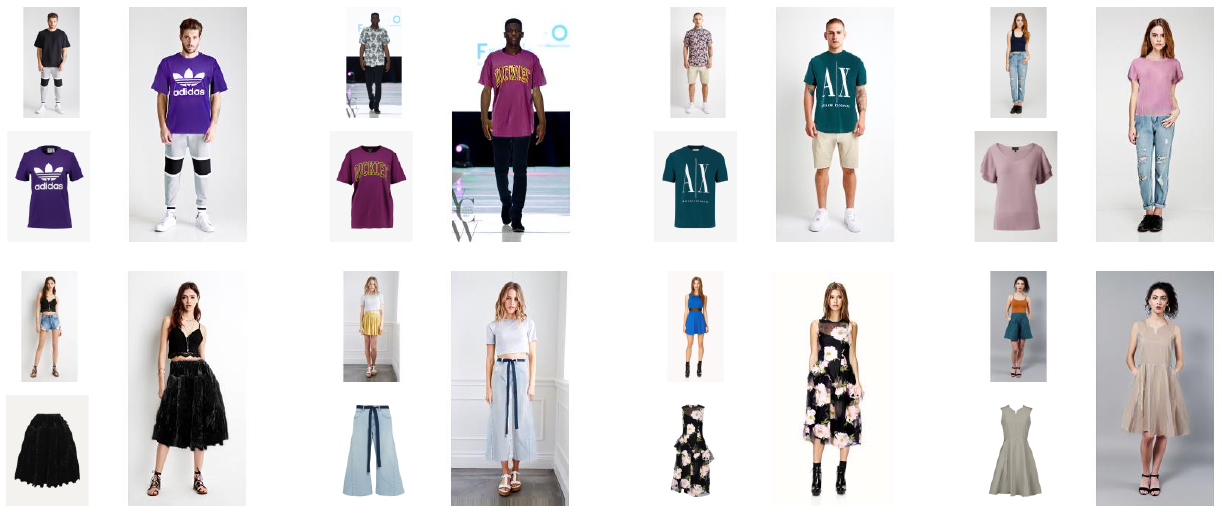}
    \caption{
    Qualitative results on SHHQ-1.0~\cite{fu2022shhq} model images with garments from DressCode~\cite{morelli2022dresscode} and VITON-HD~\cite{choi2021vitonhd}.
    }
    \label{fig:show3}
\end{figure*}

\section{Experimentation}

\subsection{Experimental Setup}
\textbf{Dataset.}
We conduct extensive experiments on two high-resolution datasets from the VITON benchmark: VITON-HD~\cite{choi2021vitonhd} and DressCode~\cite{morelli2022dresscode}. 
The VITON-HD dataset includes 13,679 pairs of images, each featuring a front-view upper-body shot of women alongside corresponding in-store garments, split into 11,647 training pairs and 2,032 testing pairs. 
The DressCode dataset is larger, with 48,392 training pairs and 5,400 testing pairs, featuring front-view full-body images of individuals with corresponding in-store garments, categorized into upper-body, lower-body, and dresses.

To train the apparel localization network, we use model images from all DressCode. We extract relevant clothing portions from the DressCode training set label maps as our ground truth, \(\mathcal{M}_{GT}\). We also integrate in-store garment images and their masks from the VITON-HD training set into \(\mathcal{M}_{GT}\).
For evaluation, we use the test sets from DressCode and VITON-HD to assess our method.

\textbf{Evaluation Metrics.}
We assess both paired and unpaired scenarios. In the paired scenario, the target human image and its corresponding garment image are used for reconstruction. In the unpaired scenario, different garment images are used for the virtual try-on experience.
To evaluate the quality of images generated in the paired scenario, we use LPIPS~\cite{zhang2018lpips} and SSIM~\cite{wangzhou2004ssim} metrics to measure resemblance to the original image. In the unpaired scenario, we use FID~\cite{heusel2017fid} and KID~\cite{binkowski2018kid} to gauge the realism and fidelity of the synthesized images.

\textbf{Implementation Details.}
We use the Adam optimizer~\cite{kingma2014adam} for the apparel localization network with a learning rate of 1e-4, halving it every 10 epochs. The network is trained for 35 epochs on two RTX 3090 GPUs. During inference, the number of clusters $K$ is set to 5.
We set the apparel-related prompt to ``clothes'' and apply our method to Stable Diffusion XL (SDXL)~\cite{podell2023sdxl}. For the inversion phase, we use an empty prompt, and for model image generation, the prompt is ``model wearing clothes.'' We conduct DDIM sampling~\cite{song2020ddim} with 50 steps and set the classifier-free guidance to 7.5.
Garment latent infusion occurs at time step \( t_1 = 40 \), and background restoration at \( t_2 = 15 \). 
The entire generation process is carried out on a single RTX 3090 GPU.

\subsection{Experimental Results}

\textbf{Quantitative Results.}
Table\,\ref{tab:result} shows the quantitative comparisons between DiffusionTrend and other methods on VITON-HD~\cite{choi2021vitonhd} and DressCode~\cite{morelli2022dresscode} test datasets. The results on DressCode of DCI-VTON~\cite{gou2023dci} and StableVITON~\cite{kim2023stableviton} contain our implementation because no usable codes are given.
Despite relying solely on pre-trained models for tasks like garment latent infusion and background restoration, our performance lags behind SOTA models. 
Past methods incur high training costs to ensure generated try-on images remain realistic and natural under various complex poses, as confirmed by evaluation metrics.
In contrast, DiffusionTrend aims to provide a resource-efficient, user-friendly tool for quickly confirming purchase intentions under simple try-on poses.
Without training the diffusion model on large datasets, our basic operations fall short in handling complex poses, leading to suboptimal performance.
As the first visual try-on model without training on diffusion models, our approach differs from traditional methods in motivation and technical approaches, making traditional try-on dataset evaluations insufficient for comprehensively measuring our method's effectiveness.
Thus, quantitative experiments serve only as a reference, while qualitative experiments will further demonstrate our superiority.

\begin{figure*}[!t]
    \centering
    \includegraphics[width=0.85\textwidth]{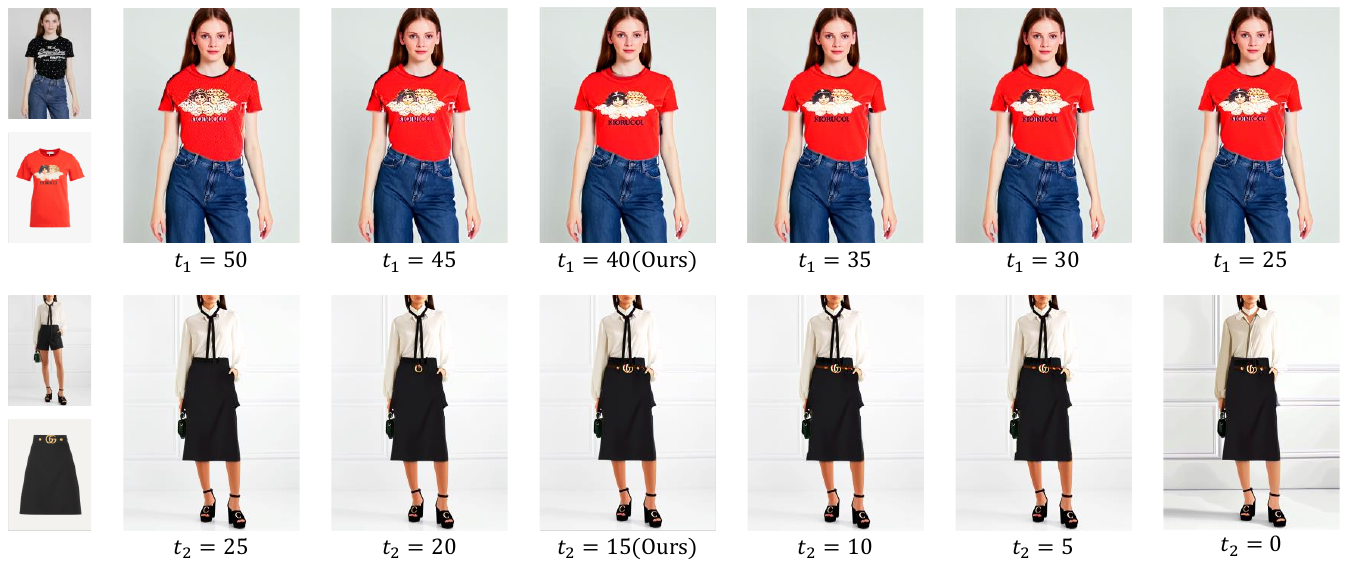}
    \caption{
    Visual ablations for $t_1$ in Garment Latent Infusion and $t_2$ in Background Restoration.
    }
    \label{fig:ablation1.}
\end{figure*}

\textbf{Qualitative Results.}
Fig.\,\ref{fig:show1} provides the qualitative comparison of DiffusionTrend with the state-of-the-art baselines
on the VITON-HD~\cite{choi2021vitonhd} and DressCode~\cite{morelli2022dresscode} datasets.
The results indicate that our DiffusionTrend performs as well as baseline methods under simple poses. 

First, most baseline methods fail to generate realistic wrinkles after applying warp techniques, simply transferring wrinkles from in-store garment images. 

Second, our approach extracts richer detail features from noise latent, allowing for more accurate detail reconstruction in complex garment patterns. 
For instance, in the first row of Fig.\,\ref{fig:show1}, the cartoon pattern on the garments reconstructed by our method more closely resembles the original image compared to other baseline methods. 
In the fourth row, while most methods erroneously reconstruct the garment as a short skirt, our approach captures the appropriate skirt length and details the metallic embellishments at the waist.

To further evaluate the generalizability of our method across datasets, we use person images from SHHQ-1.0~\cite{fu2022shhq} as the model image and garments from DressCode~\cite{morelli2022dresscode} and VITON-HD~\cite{choi2021vitonhd} for try-on. As shown in Fig.\,\ref{fig:show3}, our method produces visually coherent and naturally blended results, demonstrating stable performance and generalizability across datasets.
%

\textbf{Accuracy of the Precise Apparel Localization Network.}
To verify the effectiveness of our proposed precise apparel localization network, we conduct experiments on the test sets of the DressCode and VITON-HD datasets. 
For the DressCode dataset, we utilize label maps to extract ground truth masks, while for the VITON-HD dataset, we use image-parse-v3. 
In both cases, we extract specific colored pixels and convert them into binary masks. We then use our apparel localization network to extract clothing masks and evaluate the performance using IoU (Intersection over Union) and Dice (Dice Coefficient) metrics. Generally, a model with an IoU above 0.5 and a Dice score above 0.7 is considered applicable in the research field. 

\begin{table}[b]
\centering
\caption{
Quantitative metrics on virtual try-on datasets for validating mask prediction accuracy.}
\label{tab:acc}
\renewcommand{\arraystretch}{1.3}  
\setlength{\tabcolsep}{2mm}        

\resizebox{\linewidth}{!}{        
\begin{tabular}{l|c|c|c|c}  
\toprule
\multirow{2}{*}{Metrics} & \multicolumn{3}{c|}{DressCode}  & \multirow{2}{*}{VITON-HD} \\
\cline{2-4}
                         & upper\_body & lower\_body & dresses &                           \\
\midrule
IoU                      & 0.7213      & 0.7138      & 0.7969  & 0.7357                    \\
\hline
Dice                     & 0.8162      & 0.8038      & 0.8538  & 0.8194                    \\
\bottomrule
\end{tabular}
}
\end{table}

Our results, as presented in Table\,\ref{tab:acc}, demonstrate that our compact CNN achieves high accuracy in mask extraction tasks, with IoU scores reaching 0.7 and Dice scores exceeding 0.8. While its accuracy may lag behind SOTA segmentation models, it is more than sufficient for handling virtual try-on tasks. These results validate the effectiveness and practicality of our approach.

\subsection{Ablation Study}
\label{sec:ablation}
In this section, we examine the optimality of our method's components and experimental settings, including background restoration and the timestep for latent infusion or replacement (Sec.\,\ref{sec:reconstruction}), garment mask refinement (Sec.\,\ref{sec:mask}), garment warping and adaptive sampling termination (Sec.\,\ref{sec:reconstruction}, Sec.\,\ref{sec:termination}).

\begin{table*}[!t]
\caption{Quantitative ablations for $t_1$ in Garment Latent Infusion and $t_2$ in Background Restoration. \textbf{Bold} numbers indicate the best performance.}
\setlength{\tabcolsep}{7mm} 
\renewcommand{\arraystretch}{1.2} 
\label{tab:ablation}
\resizebox{\linewidth}{!}{ 
\begin{tabular}{c|c|c|c|c|c}
\toprule
\multicolumn{2}{c|}{Timesteps}                  & \multicolumn{4}{c}{Metrics} \\
\midrule
$t_1$                 &$t_2$                  & LPIPS$\downarrow$ & SSIM$\uparrow$ & FID$\downarrow$ & KID$\downarrow$ \\
\midrule
50                   & \multirow{6}{*}{15} & 0.0791                    & 0.9142                   & 9.8943                 & \textbf{0.43}                    \\
45                   &                     & 0.0756                    & 0.9153                   & 9.7351                  & \textbf{0.43}                    \\
40 (Ours)            &                     & 0.0720                    & 0.9172                   & \textbf{9.7040}                  & \textbf{0.43}                    \\
35                  &                     & 0.0712                    & 0.9177                   & 9.8118                  & 0.45                    \\
30                  &                     & 0.0698                    & 0.9183                    & 10.0197                 & 0.47                    \\
25                  &                     & \textbf{0.0691}                    & \textbf{0.9187}                   & 10.1865                 & 0.47                    \\
\midrule
\multirow{6}{*}{40} & 25                  & 0.0754                    & 0.9144                   & \textbf{9.2217}                  & \textbf{0.40}                    \\
                    & 20                  & 0.0737                    & 0.9160                   & 9.4248                  & 0.42                    \\
                    & 15 (Ours)            & \textbf{0.0720}                    & 0.9172                  & 9.7040                  & 0.43                    \\
                    & 10                  & 0.0724                    & \textbf{0.9173}                    & 10.0531                 & 0.46                    \\
                    & 5                  & 0.0727                    & \textbf{0.9173}                    & 10.6983                 & 0.50                  \\
                    & 0                  & 0.1676                    & 0.8527                   & 15.9559                 & 0.73     \\
\bottomrule
\end{tabular}
}
\end{table*}

\textbf{$t_1$ for Garment Latent Infusion.}
The upper part of Table\,\ref{tab:ablation} shows that incorporating clothing information too early tends to lower the LPIPS, SSIM, and FID scores. Specifically, as observed in the top row of Fig.\,\ref{fig:ablation1.}, prematurely incorporating garment information results in a deficiency of integration between the model and the garment, effectively reconstructing their respective latent representations in disparate areas without any interaction. This issue is clearly visible, as there are pronounced boundaries between the clothing and the human figure, creating an impression of disjunction rather than a cohesive unity.
On the other hand, introducing the information at a later stage impedes the development and enhancement of the garment's finer details. This is observable in the figure, where the cartoon patterns are prone to becoming indistinct, thereby compromising the overall quality of the image.

\textbf{$t_2$ for Background Restoration.}
Although the quantitative results in Table\,\ref{tab:ablation} show that the FID and KID scores are higher at $t_2=25$ and $20$, it can be observed from the second row of Fig.\,\ref{fig:ablation1.} that, performing background restoration too early can negatively impact the generation of garment details. For instance, at $t_2 = 25$, the waist's metallic embellishments are lost during the reconstruction process; at $t_2 = 20$, the situation improves slightly, but the metallic dots on either side are still missing.
In contrast, performing it too late creates a distinct boundary between the background and the foreground, resulting in unnatural outcomes, and leads to the over-rendering of details.
In the figure at $t_2 = 15$, there are small red artifacts visible around the metallic dots.

\begin{figure}[h] 
    \centering
    \includegraphics[width=0.48\textwidth]{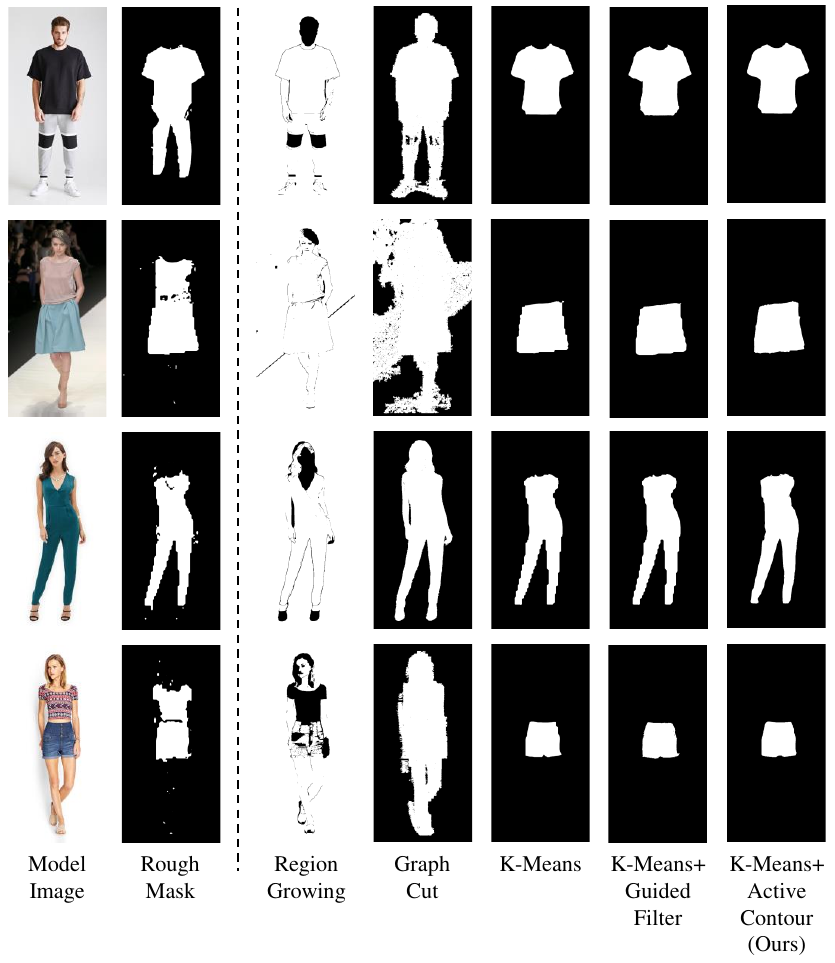}
    \caption{Visual comparison of different garment structure decomposition strategies.} 
    \label{fig:kmeans_mask}
\end{figure}

\textbf{Garment Mask Refinement Strategy.}
Initially, to determine an effective approach for separating upper and lower garments from the coarse masks, we evaluate several structural decomposition methods. Region growing tends to over-expand into background areas due to intensity fluctuations, while graph cut often includes the entire human silhouette based on global contrast, deviating from our goal of localized garment segmentation.
We observe that a basic K-means clustering can achieve a reasonably effective separation between upper and lower garments.
To further improve the segmentation precision and boundary quality, we explore two post-processing techniques: guided filtering and active contour. The former smooths edges but lacks structural modeling ability, while the latter explicitly optimizes contour continuity through energy minimization.
Considering both performance and efficiency, we adopt the combination of K-means clustering and active contour refinement as our final strategy. This setup achieves more natural garment boundaries with minimal additional overhead. Qualitative comparisons are presented in Fig.\,\ref{fig:kmeans_mask}.

\begin{figure*}[!t] 
    \centering
    \vspace{-1.0em}
    \includegraphics[width=0.95\textwidth]{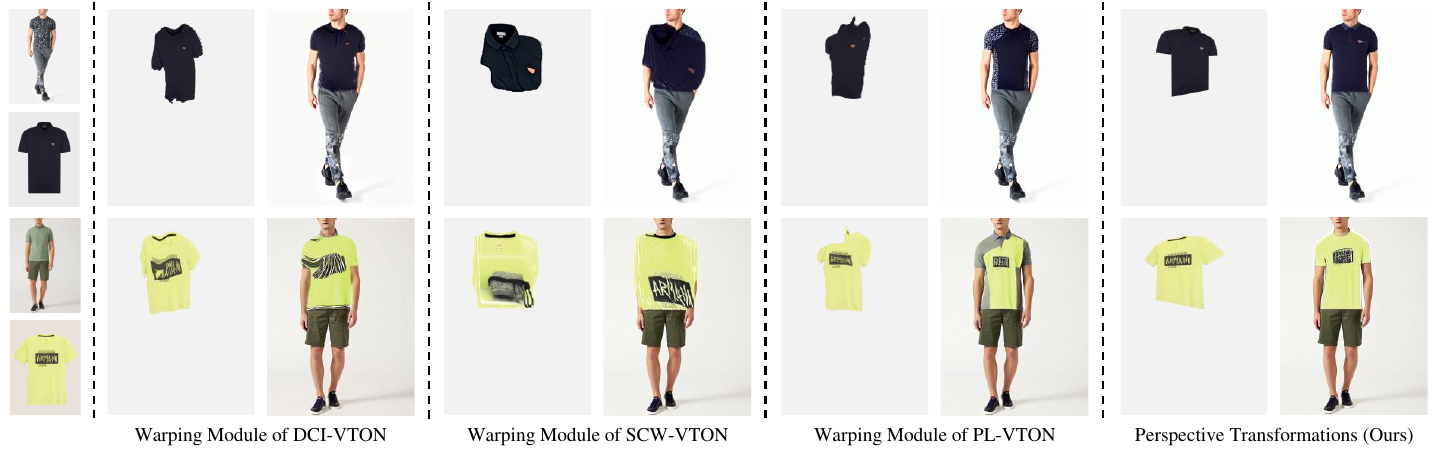}
    \caption{Comparison of results using different explicit warping modules and perspective transformations. Left: garment with warped/perspective transformation; Right: results generated using the processed garment.} 
    \label{fig:warping}
    \vspace{-1.0em}
\end{figure*}

\textbf{Garment Warping Strategy.}
To assess the feasibility of garment warping strategies, we conduct a comparative study on several representative explicit warping modules and perspective transformations. Specifically, we extract the warping modules from: (1) DCI-VTON~\cite{gou2023dci}, which represents a dense correspondence-based appearance flow approach; (2) SCW-VTON~\cite{han2024scwvton}, which introduces shape constraints into the appearance flow to improve contour alignment; (3) PL-VTON~\cite{zhang2023plvton}, which combines affine transformation with dense flow.
These approaches represent typical warping modules in diffusion-based and CNN-based virtual try-on pipelines. All of them require dedicated training and rely on structured annotations such as parsing maps and keypoints. To ensure a fair comparison, we integrate their warped garment outputs into the DiffusionTrend pipeline and compared the try-on results with those using perspective transformation. The visual comparison is presented in Fig.\,\ref{fig:warping}.

Both visual results and technical analysis reveal several shared drawbacks among these warping methods: 
(1) Unstable results with distorted edges: Outputs vary considerably across samples, often showing irregular deformations or missing garment parts after warping.
(2) Poor generalization and weak robustness: These modules, trained on specific datasets (e.g., VITON~\cite{han2017viton} or VITON-HD~\cite{choi2021vitonhd}), but exhibit clear structural failures on the DressCode~\cite{morelli2022dresscode} dataset, as shown in our visual examples.
(3) Heavy reliance on structural inputs:  Most methods require annotations such as keypoints, parsing maps, or densepose, which increases system complexity and reduces accessibility.
(4) Strong coupling with downstream try-on models: Although the warping modules are trained independently, their outputs are typically used together with try-on networks during training, leading to joint optimization where the try-on network compensates for warping errors to some extent. However, since we do not train the diffusion model, these imperfect warped results cannot be rectified by the diffusion process itself, leading to failure cases. 

Beyond trained warping modules, we also explore non-learning heuristics such as region-based pasting using parsing masks or relative keypoint-based garment stretching. While theoretically simple and training-free, these approaches also rely on structured annotations and showed inferior visual consistency.

In light of these conclusions, we retain perspective transformation as our final garment alignment strategy. Despite its simplicity, it offers consistent and controllable warping results without requiring structural supervision or additional training, making it well-suited to the training-free paradigm of DiffusionTrend.

\begin{figure*}[!t] 
    \centering
    \includegraphics[width=0.95\textwidth]{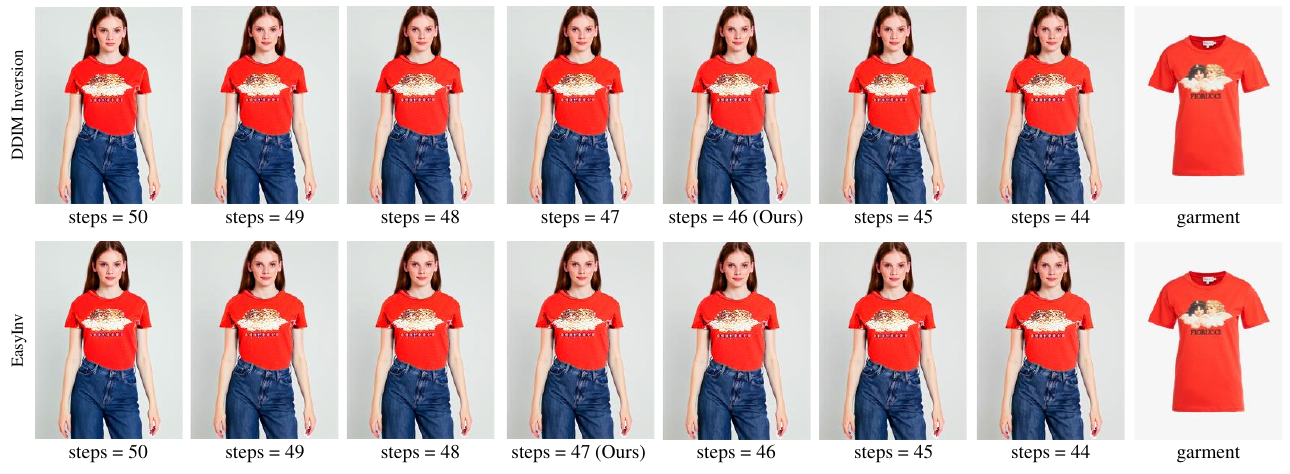}
    \caption{Effectiveness of our adaptive sampling termination across different inversion methods.} 
    \label{fig:color}
\end{figure*}

\textbf{Adaptive Sampling Termination.}
We conduct an experiment in which the number of sampling steps is gradually reduced from 50 to 44, in order to evaluate how this parameter affects the generation quality of DiffusionTrend and to verify the effectiveness of our adaptive sampling termination strategy.
As shown in Fig.\,\ref{fig:color}, varying the sampling steps leads to three major types of degradation:  
(1) Color deviation, where higher sampling steps (e.g., 50) produce oversaturated garments and backgrounds with noticeable reddish or overly bright tones;  
(2) Loss of detail, where steps below a threshold (e.g., $<45$) result in blurry textures and contours; and  
(3) Structural distortion, where overly few steps lead to background blending failures and facial deformations.

We attribute this sensitivity to two key factors. First, the training-free nature of DiffusionTrend leaves sampling steps as the only adjustable variable to balance quality and efficiency. Second, a larger number of sampling steps tends to amplify prompt influence, which can cause unwanted deviation from the appearance of the reference garment.

Our adaptive sampling termination strategy mitigates this by terminating sampling at the point where visual fidelity begins to decline. To ensure early-stage stability, this termination strategy is only activated after step $t = 45$ in our implementation.
As shown in Fig.\,\ref{fig:color}, under the standard SDXL~\cite{podell2023sdxl} + DDIM Inversion~\cite{song2020ddim} pipeline, our adaptive sampling termination strategy consistently terminates at step 46, aligning well with our empirical observation.  
We further applied the same strategy to the SDXL + EasyInv~\cite{zhang2024easyinv} setup, where the model automatically selected step 47 as the optimal stopping point. The generated results are visually consistent and stable, demonstrating the robustness and reliability of our strategy across different settings.

\section{Limitations and Future Work}
\label{sec:limitation}

\begin{figure}[!t] 
    \centering
    \includegraphics[width=0.45\textwidth]{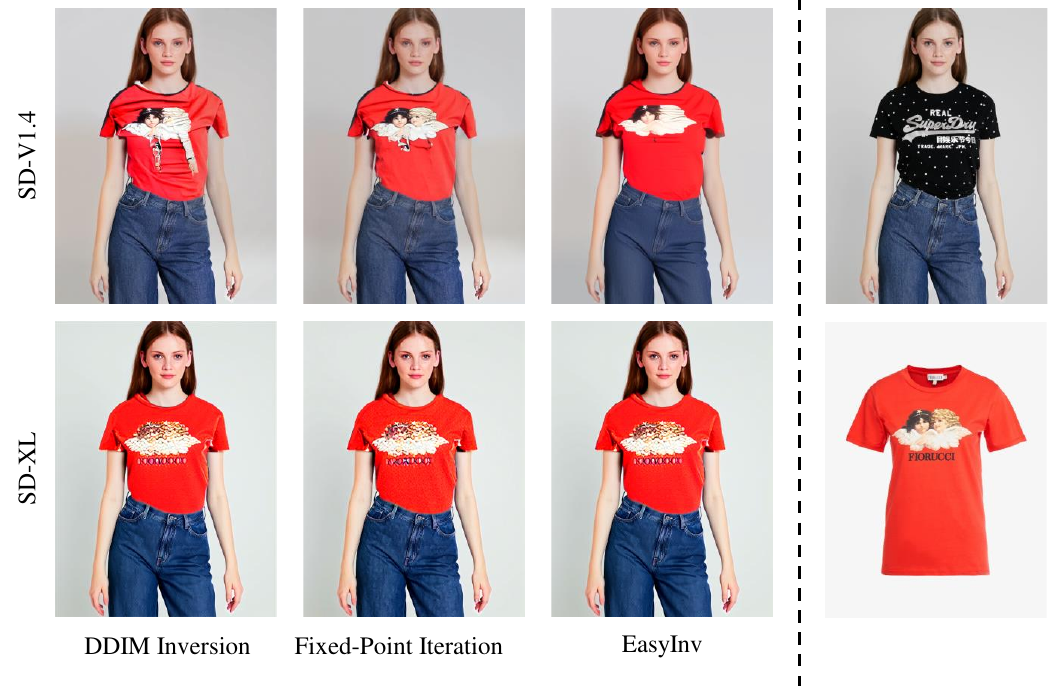}
    \caption{Improved inversion results achieved by SD-XL and advanced inversion methods.} 
    \label{fig:inversion}
\end{figure}

Our DiffusionTrend model encounters several challenges: 

\textbf{Reliance on Inversion Quality.}
Our proposed DiffusionTrend's performance is indeed influenced by the quality of DDIM inversion results. However, we find that the evolution of more powerful diffusion models is likely to yield better results even with the same inversion method, As demonstrated in Fig.\,\ref{fig:inversion}. 
Moreover, ongoing advancements in inversion methods, such as NULL-Text Inversion~\cite{mokady2023nulltextinversion}, ReNoise~\cite{garibi2024renoise}, Fixed-Point Iteration~\cite{pan2023fixedpoint}, and EasyInv~\cite{zhang2024easyinv}, provide promising directions for further improvements.

Furthermore, our framework is inherently compatible with ongoing inversion advances and can directly benefit from them without any architectural changes. We thus view current inversion limitations not as constraints, but as evolving components of an improving ecosystem.
We believe that the emergence of these improved diffusion models and inversion techniques will significantly expand the application space of our training-free try-on paradigm.

\textbf{Challenges in Adapting to Complex Poses.}
It encounters difficulties in generating complex poses, especially in rendering body parts that are not visible in the original model image, due to limitations inherent in the pre-trained model.
For instance, if the original model is depicted wearing a long-sleeved shirt and the target garment is a short-sleeved one, the model is unable to convincingly render the exposed arms—a challenge commonly faced by most virtual try-on models.
These limitations suggest that future research should prioritize better recovery of fine-grained details in clothing and strengthen the model's ability to generate unseen body parts. 

Moreover, due to the limitations of perspective transformations, DiffusionTrend encounters difficulties when dealing with complex model poses, particularly in accurately aligning garments with intricate body movements or occluded regions. 
While regular poses are generally sufficient for consumers to make purchase decisions by providing a clear view of garment fit and style, further exploration into addressing these challenges and refining the quality of background restoration are areas that merit deeper investigation.

\textbf{Inference Overhead and Optimization Prospects.}
DiffusionTrend requires additional time during inference due to the DDIM inversion process, which involves a similar number of steps as DDIM sampling. This time overhead mainly occurs during the inference process and is a potential drawback. However, it can be mitigated through quantization (\emph{e.g.}, Q-Diffusion~\cite{li2023qdiffusion}, NDTC~\cite{shang2023post}) and pruning (\emph{e.g.}, LD-Pruner~\cite{castells2024ld}). While this paper focuses on reducing training overhead, combining these techniques to optimize inference speed will be an important direction for future research.

Despite these shortcomings, DiffusionTrend offers a low-cost, lightweight paradigm for the virtual try-on field that circumvents extensive diffusion model training. Optimism remains high that this approach will continue to evolve as the field advances. We respectfully ask the academic community to recognize the value of this exploratory work and extend the necessary patience and support for further development.

\section{Conclusion}
In this paper, we have introduced DiffusionTrend, a novel try-on methodology that forgoes the need for training diffusion models, thereby offering straightforward, conventional pose virtual try-on services with minimal computational demands.
Capitalizing on sophisticated diffusion models, DiffusionTrend harnesses latents brimming with prior information to encapsulate the nuances of garment details. 
Throughout the diffusion denoising process, these details are effortlessly merged into the model image generation, expertly directed by a precise garment mask generated by a lightweight and compact CNN.
Differing from other approaches, DiffusionTrend sidesteps the necessity for labor-intensive training of diffusion models on extensive datasets. 
It also dispenses with the need for various types of user-unfriendly model inputs. 
Our experiments demonstrate that, despite lower metric performance, DiffusionTrend delivers a visually convincing virtual try-on experience, all while maintaining the quality and richness of fashion presentation.

\bibliographystyle{IEEEtran}
\bibliography{main}

\begin{IEEEbiography}[{\includegraphics[width=1in,height=1.25in,clip,keepaspectratio]{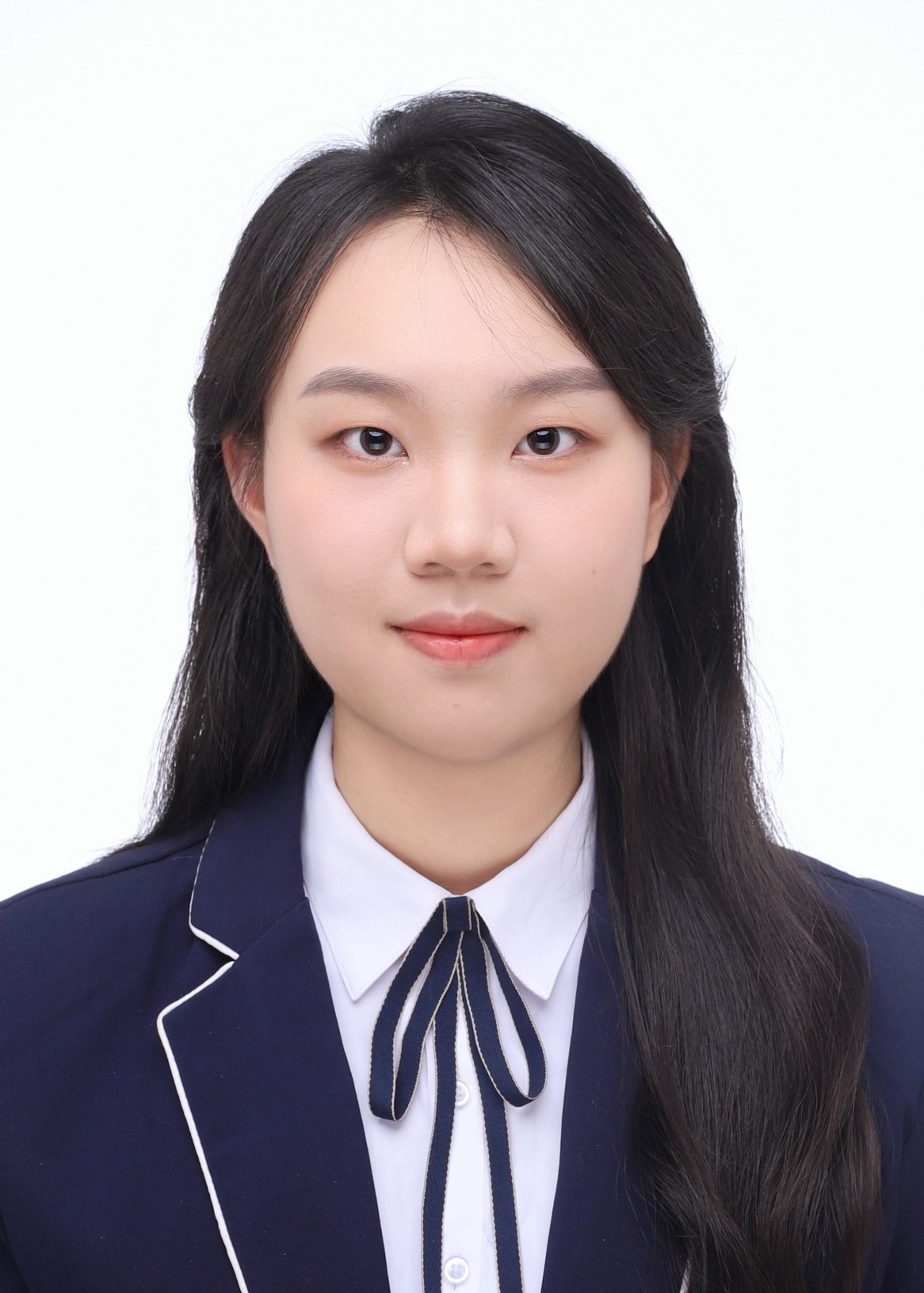}}]{Wengyi Zhan}  completed her undergraduate studies and obtained the Bachelor's degree in Intelligence Science and Technology from Xiamen University, Xiamen, China, in 2023. 

She is currently a master's student at the School of Informatics, Xiamen University, specializing in Intelligence Science and Technology. Her current research interest focuses on diffusion models for consistent video/image generation.
\end{IEEEbiography}

\begin{IEEEbiography}[{\includegraphics[width=1in,height=1.25in,clip,keepaspectratio]{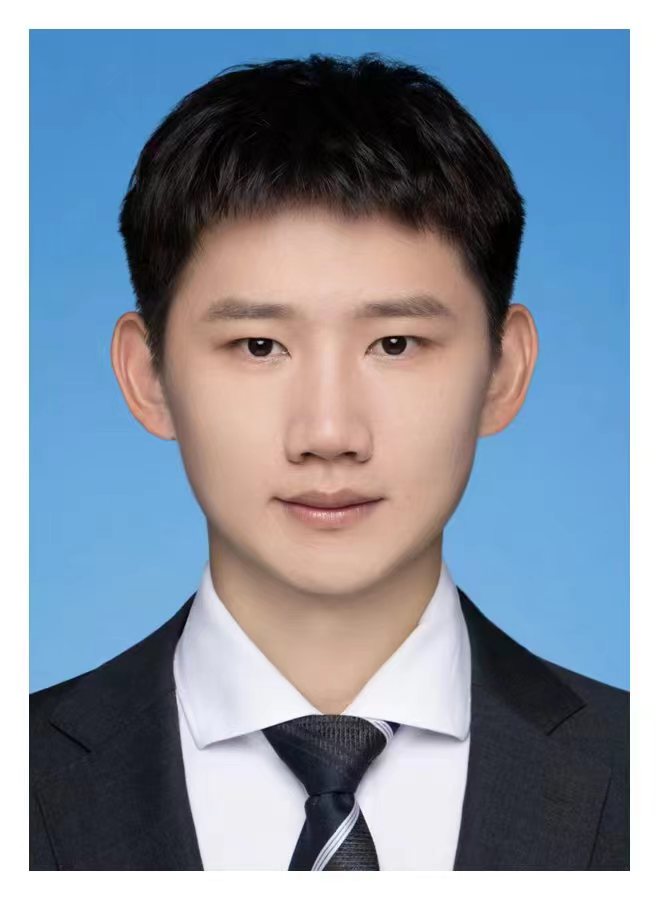}}]{Mingbao Lin} finished his M.S.-Ph.D. study and obtained the Ph.D. degree in intelligence science and technology from Xiamen University, Xiamen, China, in 2022. Earlier, he received the B.S. degree from Fuzhou University, Fuzhou, China, in 2016.

He is currently a research scientist with the Skywork AI, Singapore, and also an adjunct industry supervisor with Xiamen University. His current research interest is to develop low-latency multimodal interaction system, such as text, audio, image, video, \emph{etc}.
\end{IEEEbiography}

\begin{IEEEbiography}[{\includegraphics[width=1in,height=1.25in,clip,keepaspectratio]{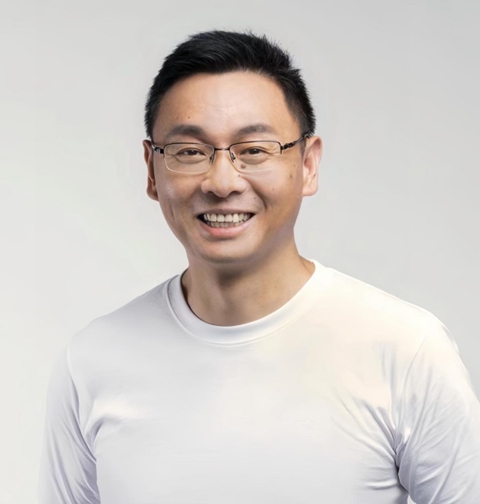}}]{Shuicheng Yan}
(Fellow, IEEE)  is currently the Managing Director of Kunlun 2050 Research and Chief Scientist of Kunlun Tech \& Skywork AI, and the former Group Chief Scientist of Sea. Prof. Yan Shuicheng is a Fellow of Singapore's Academy of Engineering, AAAI, ACM, IEEE, and IAPR. His research areas include computer vision, machine learning, and multimedia analysis. Till now, Prof Yan has published over 800 papers at top international journals and conferences, with an H-index of 140+. He has also been named among the annual World's Highly Cited Researchers nine times. Prof. Yan's team received ten-time winners or honorable-mention prizes at two core competitions, Pascal VOC and ImageNet (ILSVRC), deemed the “World Cup” in the computer vision community. Besides, his team won more than ten best papers and best student paper awards, particularly a grand slam at the ACM Multimedia, the top-tiered conference in multimedia, including the Best Paper Awards thrice, Best Student Paper Awards twice, and Best Demo Award once.
\end{IEEEbiography}

\begin{IEEEbiography}[{\includegraphics[width=1in,height=1.25in,clip,keepaspectratio]{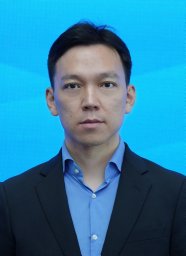}}]{Rongrong Ji}
(Senior Member, IEEE) is a Nanqiang Distinguished Professor at Xiamen University, the  Director of the Office of Science and Technology at Xiamen University, and the Director of Media Analytics and Computing Lab. He was awarded as the National Science Foundation for Excellent Young Scholars (2014), the National Ten Thousand Plan for Young Top Talents (2017), and the National Science Foundation for Distinguished Young Scholars (2020). His research falls in the field of computer vision, multimedia analysis, and machine learning. He has published 50+ papers in ACM/IEEE Transactions, including TPAMI and IJCV, and 100+ full papers on top-tier conferences, such as CVPR and NeurIPS. His publications have got over 20K citations in Google Scholar. He was the recipient of the Best Paper Award of ACM Multimedia 2011. He has served as Area Chairs in top-tier conferences such as CVPR and ACM Multimedia. He is also an Advisory Member for Artificial Intelligence Construction in the Electronic Information Education Committee of the National Ministry of Education.
\end{IEEEbiography}


\vfill

\end{document}